\documentclass[sigconf,authorversion=true, nonacm=true]{acmart}

\setcopyright{none}

\usepackage{booktabs} 
\usepackage{mathtools}
\usepackage{float}
\usepackage{graphicx}
\usepackage{subcaption}
\usepackage{multirow}
\usepackage{rotating}
\usepackage{array}
\usepackage{gensymb}
\usepackage{pdfpages}
\usepackage{multirow}
\usepackage{makecell}
\usepackage{wrapfig}
\usepackage{algorithm}
\usepackage[noend]{algpseudocode}
\usepackage{tikz}
\usetikzlibrary[topaths]
\usetikzlibrary{arrows,decorations.markings,arrows.meta}
\usetikzlibrary{positioning}
\tikzset{
    position/.style args={#1:#2 from #3}{
        at=(#3.#1), anchor=#1+180, shift=(#1:#2)
    }
}

\newcommand{\mc}[1]{\mathcal{#1}}
\newcount\mycount
\newcount\mynode

\newcommand{\mycomment}[1]{}

\makeatletter
\newsavebox\myboxA
\newsavebox\myboxB
\newlength\mylenA

\def\vec#1{\mathchoice{\mbox{\boldmath$\displaystyle#1$}}
{\mbox{\boldmath$\textstyle#1$}}
{\mbox{\boldmath$\scriptstyle#1$}}
{\mbox{\boldmath$\scriptscriptstyle#1$}}}
\makeatother

\makeatletter
\newcommand*{\bigominus}{\DOTSB\bigominus@\slimits@}

\newcommand{\bigominus@}{\mathop{\mathpalette\bigominus@@\relax}}

\newcommand{\bigominus@@}[2]{%
  \vcenter{\hbox{%
    \sbox\z@{$\m@th#1\bigoplus$}%
    \resizebox{\wd\z@}{!}{$\m@th#1\bm{\ominus}$}%
  }}%
}
\makeatother

\algblock{ParFor}{EndParFor}
\algnewcommand\algorithmicparfor{\textbf{parfor}}
\algnewcommand\algorithmicpardo{\textbf{do}}
\algnewcommand\algorithmicendparfor{\textbf{end\ parfor}}
\algrenewtext{ParFor}[1]{\algorithmicparfor\ #1\ \algorithmicpardo}
\algrenewtext{EndParFor}{\algorithmicendparfor}

\makeatletter
\ifthenelse{\equal{\ALG@noend}{t}}%
  {\algtext*{EndParFor}}
  {}%
\makeatother

\begin{document}

\title{GPU-Accelerated Parallel Gene-pool Optimal Mixing in a Gray-Box Optimization Setting}

\author{Anton Bouter}
\affiliation{%
  \institution{Centrum Wiskunde \& Informatica}
  \city{Amsterdam}
  \country{The Netherlands}
}
\email{Anton.Bouter@cwi.nl}

\author{Peter A.N. Bosman}
\affiliation{%
  \institution{Centrum Wiskunde \& Informatica}
  \city{Amsterdam}
  \country{The Netherlands}
}
\email{Peter.Bosman@cwi.nl}

\begin{abstract}
In a Gray-Box Optimization (GBO) setting that allows for partial evaluations, the fitness of an individual can be updated efficiently after a subset of its variables has been modified. This enables more efficient evolutionary optimization with the Gene-pool Optimal Mixing Evolutionary Algorithm (GOMEA) due to its key strength: Gene-pool Optimal Mixing (GOM). For each solution, GOM performs variation for many (small) sets of variables. To improve efficiency even further, parallel computing can be leveraged. For EAs, typically, this comprises population-wise parallelization. However, unless population sizes are large, this offers limited gains. For large GBO problems, parallelizing GOM-based variation holds greater speed-up potential, regardless of population size. However, this potential cannot be directly exploited because of dependencies between variables. We show how graph coloring can be used to group sets of variables that can undergo variation in parallel without violating dependencies. We test the performance of a CUDA implementation of parallel GOM on a Graphics Processing Unit (GPU) for the Max-Cut problem, a well-known problem for which the dependency structure can be controlled. We find that, for sufficiently large graphs with limited connectivity, finding high-quality solutions can be achieved up to 100 times faster, showcasing the great potential of our approach.
\end{abstract}

\keywords{Parallel Computing, CUDA, Gray-Box Optimization, Optimal Mixing}

\ccsdesc[500]{Mathematics of computing~Evolutionary algorithms}

\maketitle

\section{Introduction}
Evolutionary computation has been used to tackle many real-world problems in the past, including engineering problems \cite{dasgupta2013evolutionary}, vehicle routing \cite{jozefowiez2008multi,jabir2015multi}, and even the treatment of cancer \cite{cabrera2014multi,luong2018application}.
Though Evolutionary Algorithms (EAs) are historically most often considered as Black-Box Optimization (BBO) algorithms, many real-world problems exist where some domain knowledge about the problem, or even the entire problem definition, is known, eliminating the necessity of tackling the problem in a BBO setting.
Instead, the problem can then be optimized in a Gray-Box Optimization (GBO) setting, where (a limited amount of) domain knowledge is used to improve the effectiveness or the efficiency of the optimization algorithm.
Having such information available does not mean that the problem is not complex and can be solved using simple heuristics or convex optimization techniques.
Acknowledging this, EAs have, for example, previously used custom recombination operators \cite{deb2016breaking}, problem decompositions \cite{chicano2017optimizing}, or partial evaluations \cite{bouter2021achieving}, to achieve great success where other optimization methods failed.
In this paper, we specifically consider a GBO setting where partial evaluations, by which we mean that evaluating the change in fitness after only a few variables have changed, can be done (proportionally) more efficiently than when a full evaluation is performed.

For the optimization of many real-world problems, it is furthermore important that the total computation time is minimized, e.g., in the case where a patient is waiting to be treated for cancer \cite{luong2018application}.
For this reason, and because EAs naturally lend themselves to parallelization due to their population-based approach, meaning that the entire population can be evaluated in parallel, much successful research has been done into the application of parallel computing techniques and EAs \cite{sudholt2015parallel,alba2006parallel}.
More recently, much of this research has been focused on large-scale parallelization using Graphics Processing Units (GPUs) \cite{wong2005parallel,li2017speeding}, because the clock frequencies of Central Processing Units (CPUs) are nearing their physical limits \cite{asanovic2006landscape,brodtkorb2010state}, meaning that it is more effective to increase the number of parallel cores rather than increasing the frequency of each of these cores.

In comparison to parallelization on a number of CPUs, the application of GPUs is generally required to be much more fine-grained, however, because of the thousands of cores present in modern GPUs, and the \textit{Single Instruction, Multiple Data} (SIMD) computational model used, which requires groups of threads on the GPU to perform the same (computational) instructions simultaneously.
Therefore, unless population sizes are vast, only evaluating the population in parallel will not have a high enough degree of parallelizability to fully utilize the computational resources of a GPU.

The Gene-pool Optimal Mixing Evolutionary Algorithm (GOMEA) \cite{thierens2011optimal} has excellent synergy with a GBO setting, because the variation steps of GOMEA, performed with the Gene-pool Optimal Mixing (GOM) variation operator, are applied to subsets of variables, and only accepted if they do not decrease the fitness of the parent.
Moreover, often, many relatively small such subsets are used.
In a GBO setting, such modifications to subsets of variables can be efficiently evaluated using partial evaluations.
Such a GBO setting can be further exploited by using the fact that separate steps in GOM, which consider different subsets, are independent when the respective sets of variables that are modified are mutually independent.
Therefore, if such dependence information is known, a higher degree of parallelizability can be achieved by performing conditionally independent partial evaluations in parallel.
This means that, for each solution in the population, a large number of variation steps can be evaluated in parallel, as long as these subsets of variables are conditionally independent.
This was previously showcased in the domain of continuous optimization and applied to the real-world problem of deformable image registration, where these independent subsets could be manually identified and hard-coded a priori \cite{bouter2021accelerated}. 

In this paper, we introduce the parallel application of GOM in the discrete domain.
Moreover, we introduce a general method using graph coloring to identify conditionally independent subsets of variables, removing the need to hard-code these for a specific problem instance a priori.
Such conditionally independent subsets of variables can be found by applying graph coloring to the Variable Interaction Graph (VIG) \cite{tintos2015partition}, i.e., the graph that describes problem variables as vertices and mutual dependencies between pairs of problem variables as edges, which can be derived from any problem to which partial evaluations can be applied.
We create a CUDA \cite{nvidia2010programming} implementation of parallel GOM, and apply it to the well-known NP-hard Max-Cut problem \cite{karp1972reducibility}, because the graph describing a MaxCut instance directly translates to the VIG, making it a problem with an easily controllable dependency structure.
However, everything we described here that is needed to make the CUDA implementation work, may be applied to any problem that allows for a GBO setting where partial evaluations can be applied.
We analyze the benefit of parallel GOM for graphs with different structures, and compare it to the original non-parallelized GOMEA.

\section{Gray-Box Optimization}

\subsection{Decomposable Fitness Functions}
We define the (discrete) optimization function $f(\vec{x}) : \mathbb{Z}^\ell \rightarrow \mathbb{R}$ as the optimization function of interest that is subject to maximization.
This optimization function maps a solution $\vec{x} = [x_1,x_2,\dots,x_{\ell}]$, i.e., an instantiation of the set of all $\ell$ problem variables $\vec{X} = [X_1,X_2,\dots,X_{\ell}]$, to a fitness value $f(\vec{x}) \in \mathbb{R}$.
The set of problem variables is indexed through $\vec{\mc{I}} = [1,2,\dots,\ell]$.

In this paper, we specifically consider a GBO setting that allows for partial evaluations, previously defined for a continuous optimization setting \cite{bouter2018large}.
This means that the fitness of a solution can be efficiently updated after a modification to a small number of variables has been made.
For partial evaluations to be performed, it must be known (from domain knowledge or otherwise) how the fitness function is constructed from any number of subfunctions.
For this purpose, we define the set of $q$ subfunctions $\vec{F} = \{f_1,f_2,\dots,f_q\}$ that compose the fitness function $f$.
Each subfunction $f_i \in \vec{F}$ is a function of a subset of problem variables of $\vec{x}$, where the indices of $\vec{x}$ that this subfunction is restricted to, are defined by $\varmathbb{I}_i$.
We use the notation $\vec{x}_{\vec{Y}}$ with $Y \subseteq \mc{I}$ to denote the subset of variables of $\vec{x}$ restricted to the indices in $Y$, i.e., $\vec{x}_{\vec{Y}} = [x_{Y_1}, x_{Y_2}, \dots, x_{Y_{|Y|}}]$ with $\vec{Y} = [Y_1,Y_2,\dots,Y_{|Y|}]$.
The set $\varmathbb{I} = [\varmathbb{I}_1,\varmathbb{I}_2,\dots,\varmathbb{I}_q]$ is given by the problem definition, and determines which variables are required as input for each of the subfunctions.
Consequently, each subfunction $f_i$, defined as $f_i : \mathbb{Z}^{|\varmathbb{I}_i|} \rightarrow \mathbb{R}$, is assumed to be non-separable, and is furthermore treated as a black box.
The GBO fitness function $f$ is then defined as an aggregation over all subfunctions, as follows:

\vspace*{-3mm}
\begin{small}
\begin{align}
\label{eq:gbo}
f(x) &= g\left( f_1(\vec{x}_{\varmathbb{I}_1}) \oplus f_2(\vec{x}_{\varmathbb{I}_2}) \oplus \dots \oplus f_q(\vec{x}_{\varmathbb{I}_q}) \right),
\end{align}
\end{small}

with $\oplus$ a binary commutative operator that has a known inverse $\ominus$, e.g., addition or multiplication, and $g : \mathbb{R} \rightarrow \mathbb{R}$, named the mapping function, any (possibly non-linear) function aggregating the output of all subfunctions to the domain of the fitness function.

Note that the domain of the output of each subfunction, and that of the input of the mapping function $g$, are not required to be the continuous domain $\mathbb{R}$.
Instead, it could be any (possibly high-dimensional) domain $\mathbb{D}^{\kappa}$, or a product of different domains, as long as the output domain of each subfunction is identical to the domain of the input of the mapping function $g$, , i.e., $f_i : \mathbb{Z}^{\ell} \rightarrow \mathbb{D}^{\kappa}$ and $g : \mathbb{D}^{\kappa} \rightarrow \mathbb{R}$.
In this paper, for the sake of simplicity, and the fact that it applies to most (real-world) GBO problems, we restrict the definition of GBO to the domain where the output of each subfunction is in the continuous domain.

\subsection{Partial Evaluations}
After the modification of a variable $x_i$ of a solution $\vec{x}$, the fitness of the modified solution $\vec{x'}$ can be efficiently computed through a partial evaluation.
This partial evaluation requires the computation of all subfunctions $f_j$ that require $x_i$ as an input variable, i.e., all $f_j$ for which $i \in \varmathbb{I}_j$.
If the mapping function $g$ is the identity function, updating the fitness value of the solution is done by subtracting (more generally, using the $\ominus$ operator) the previous contribution of the subfunction to the fitness, and adding (more generally, using the $\oplus$ operator) the current contribution of the subfunction to the fitness.
If $g$ is not the identity function, it is required to keep track of the sum of all subfunctions, i.e., $f_1(\vec{x}_{\varmathbb{I}_1}) \oplus f_2(\vec{x}_{\varmathbb{I}_2}) \oplus \dots \oplus f_q(\vec{x}_{\varmathbb{I}_q})$, for each individual in the population.
The subtractions and additions of subfunctions are then applied to this sum, and used as input for the mapping function to find the fitness value.

\subsection{Variable Interaction Graph}
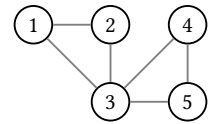
\begin{figure}[h]
\scalebox{0.5}{
\begin{tikzpicture}[transform shape]
    \node[draw,circle,minimum size=1cm,line width=0.5mm] (N1) {\Huge 1};
	\node[draw,circle,minimum size=1cm,line width=0.5mm,right=of N1] (N2) {\Huge 2};
    \node[draw,circle,minimum size=1cm,line width=0.5mm,below=of N2] (N3) {\Huge 3};
    \node[draw,circle,minimum size=1cm,line width=0.5mm,right=of N2] (N4) {\Huge 4};
    \node[draw,circle,minimum size=1cm,line width=0.5mm,below=of N4] (N5) {\Huge 5};
        
    \draw[gray,line width=0.5mm] (N1) -- (N2);
    \draw[gray,line width=0.5mm] (N1) -- (N3);
	\draw[gray,line width=0.5mm] (N2) -- (N3);
	\draw[gray,line width=0.5mm] (N3) -- (N4);
	\draw[gray,line width=0.5mm] (N3) -- (N5);
	\draw[gray,line width=0.5mm] (N4) -- (N5);
\end{tikzpicture}
}
\vspace*{-3mm}
\caption{Example of a VIG of a 5-dimensional problem.}
\label{fig:vig}
\end{figure}

The structure of (variable interactions in) an optimization problem can be captured in the VIG \cite{tintos2015partition}.
This graph $\texttt{VIG} = (\vec{V}_{\texttt{VIG}},\vec{E}_{\texttt{VIG}})$ consists of a set of vertices $\vec{V}_{\texttt{VIG}}$, one for each problem variable, and a set of edges $\vec{E}_{\texttt{VIG}}$.
Each edge $(u,v) \in \vec{E}_{\texttt{VIG}}$ denotes that variables $X_u$ and $X_v$ are dependent, i.e., a subfunction $f_i \in \vec{F}$ exists that requires both $X_u$ and $X_v$ as input.
An example of a VIG of a 5-dimensional problem is displayed in Figure \ref{fig:vig}.

\section{General Purpose Graphics Processing Units}
Due to the fact that GPUs contain a large number of computational units, and the fact that CPU cores are close to reaching their physical limits \cite{asanovic2006landscape,brodtkorb2010state}, GPUs are well known to be among the state of the art for high-performance computing, accelerating and enabling wide-spread use of deep learning.
Also in the field of evolutionary computation, the large-scale parallel computing power of GPUs has previously been used \cite{wong2005parallel,li2017speeding}.

In this paper, we use NVIDIA GPUs, for which code is developed in CUDA.
These GPUs consist of a large number, up to approximately 10,000 in modern GPUs, of computational units, named CUDA cores.
Parallel computing on NVIDIA GPUs is done with the Single Instruction/Multiple Data (SIMD) computational model, which limits the flexibility of the computations compared to a regular, serial computational model.
Executing computations on the GPU is done through kernels, which are functions that are executed in parallel for a large number of threads organized in a grid.
Within such a grid, threads are distributed into thread blocks, for which the size must be specified in the CUDA code.
Each of these threads execute the exact same code, but may access different data by using the thread ID, i.e., the location of the thread in the grid.

It is beyond the scope of this paper to provide more details of GPU architectures and computation models for which we refer the interested reader to relevant literature \cite{nvidia2010programming,nvidia2017tesla}.

\section{Gene-pool Optimal Mixing Evolutionary Algorithm}
The Gene-pool Optimal Mixing Evolutionary Algorithm (GOMEA) \cite{thierens2011optimal} is a Model-Based Evolutionary Algorithm (MBEA), of which the main strength is the Gene-pool Optimal Mixing (GOM) variation operator.
This variation operator uses an explicit linkage model, describing interactions between problem variables in terms of groups of variables called linkage sets, to perform crossover with groups of highly dependent variables, and only accepting variation operations that do not decrease the fitness of the solution.
As such, dependencies in the optimization problem are exploited, because important building blocks are not disrupted, yet mixed well.

A population $\mc{P}$ of size $n$ is maintained by GOMEA, which is typically initialized uniformly at random, although problem-specific information can be leveraged here as well.
During each generation of GOMEA, GOM is applied to each individual in the population, for each linkage set in the linkage model.
Pseudo-code for GOMEA is displayed in Algorithm \ref{alg:gomea}.

\begin{small}
\begin{algorithm}
\caption{GOMEA}\label{alg:gomea}
\begin{algorithmic}[1]
\Procedure{$\texttt{GOMEA}$}{$n$}
\State $\mc{P} \gets \texttt{InitializePopulation}(n)$
\While { $\textbf{not}\;\texttt{TerminationCriterionSatisfied}()$ }
  \State $\mc{F} \gets \texttt{LearnLinkageModel}(\mc{P})$ 
  \State $\mc{O} \gets \emptyset$ 
  \For {$\vec{x} \in \mc{P}$} \Comment{Iterate over solutions}
  	\State $\vec{o} \gets \texttt{clone}(\vec{x})$
  	\For{$\mc{F}_i \in \mc{F}$} \Comment{Random order}
  		\State $\vec{d} \gets \texttt{SelectDonorFromPopulation}(\mc{P},\mc{F}_i)$
  		\State $\vec{o} \gets \texttt{GOM}(\vec{o},\vec{d},\mc{F}_i)$
  	\EndFor
  	\State $\vec{o} \gets \texttt{ApplyForcedImprovement}(\vec{o})$
  	\State $\mc{O} \gets \mc{O} \cup \vec{o}$
  \EndFor
  \State $\mc{P} \gets \mc{O}$
\EndWhile
\EndProcedure
\end{algorithmic}
\end{algorithm}
\end{small}

\begin{small}
\begin{algorithm}
\caption{Gene-pool Optimal Mixing}\label{alg:gom}
\begin{algorithmic}[1]
\Procedure{$\texttt{GOM}$}{$\vec{o},\vec{d},\mc{F}_j$}
\State $\vec{o}' \gets \vec{d}[\mc{F}_j]$ \Comment{Variables of $\vec{d}$ restricted to indices in $\mc{F}_j$}
\State $\Delta f_o' \gets \texttt{PartialEvaluation}(\vec{o}',\vec{o},f_x,\mc{F}_j)$
\State $f_o' \leftarrow f_o + \Delta f_o'$
\State $\texttt{acceptVariation} \gets \textbf{False}$
\If { $f_o' > f_o$ }
	\State $\texttt{acceptVariation} \gets \textbf{True}$
\ElsIf{$f_o' == f_o\;\textbf{and not}\;\texttt{EqualGenotype}(\vec{o},\vec{x}^{\texttt{elitist}})$}
	\State $\texttt{acceptVariation} \gets \textbf{True}$
\EndIf
\If{ $\texttt{acceptVariation}$ }
	\State $f_o \leftarrow f_o + \Delta f_o'$ \Comment{Update fitness of $\vec{o}$}
	\State $\vec{o}[\mc{F}_j] \gets \vec{o}'$ \Comment{Update genotype for indices in $\mc{F}_j$}
\EndIf
\EndProcedure
\end{algorithmic}
\end{algorithm}
\end{small}

In Algorithm \ref{alg:gomea}, the $\texttt{InitializePopulation}$ function initializes a population $\mc{P}$ of size $n$ uniformly at random.
At the start of each generation, a linkage model is learned based on the population $\mc{P}$.
In a GBO setting, however, it may be possible to learn a linkage model offline, before optimization, based on domain knowledge, and keeping it constant throughout a run of GOMEA.
In this setting, the linkage model learning procedure at the start of each generation would therefore be omitted.
The $\texttt{SelectDonorFromPopulation}$ function randomly selects a donor from the population for which the genotype, restricted to the problem variables in the respective linkage set, are not equal to that of the parent solution $\vec{o}$.
If no such donor is present in the population, GOM is not performed for the current linkage set, and continues to the next iteration.
The $\texttt{ApplyForcedImprovement}$ function applies the Forced Improvement (FI) \cite{bosman2012linkage} procedure to solutions for which no variation step of GOM in the current generation was accepted, or for which the fitness has not improved for a certain number of generations.
During the FI procedure, the respective solution $\vec{o}$ undergoes variation with GOM for each linkage model.
However, instead of randomly selecting a donor from the population, the elitist solution is used as the donor.
Moreover, if any such variation step of GOM improves the fitness of $\vec{o}$, the FI procedure is terminated.
If instead, at the end of the FI procedure, no improvement was made to $\vec{o}$, a copy of the elitist solution $\vec{x}^{\texttt{elitist}}$ takes its place in the population.

\subsection{Linkage Model}
\label{subsec:linkagemodel}
The linkage model used by GOMEA is a Family Of Subsets (FOS) $\vec{\mc{F}} = \{\mc{F}_1,\mc{F}_2,\dots,\mc{F}_m\}$, with $\mc{F}_i \subseteq \vec{\mc{I}}$ for each $\mc{F}_i \in \vec{\mc{F}}$.
Each element $\mc{F}_i$ of the linkage model, named a linkage set, defines a subset of variables that is considered to be strongly dependent.
Note that, each FOS used to describe a linkage model should be complete, i.e., contain each problem variable in at least one linkage set.

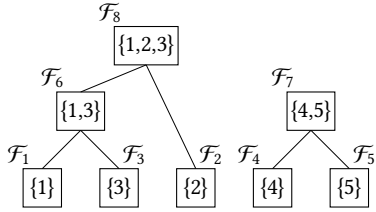
\begin{figure}[h]
\vspace*{-2mm}
\centering
\resizebox{0.6\linewidth}{!}{
\begin{tikzpicture}
\tikzset{
roundnode/.style={
  rectangle,
  outer sep=0cm,
  align=center,
  draw=black,
  fill=white,
  thick,
  minimum size=10mm
  }
}

\node[roundnode]      (1)                    {\Huge \{1\}};
\node[roundnode]      (3)       [right=of 1] {\Huge \{3\}};
\node[roundnode]      (2)       [right=of 3] {\Huge \{2\}};
\node[roundnode]      (4)       [right=of 2] {\Huge \{4\}};
\node[roundnode]      (5)       [right=of 4] {\Huge \{5\}};
\node[roundnode]      (13)      [above=of 1,xshift=1cm] {\Huge \{1,3\}};
\node[roundnode]      (123)     [position=45:1cm from 13] {\Huge \{1,2,3\}};
\node[roundnode]      (45)      [above=of 4,xshift=1cm] {\Huge \{4,5\}};

\node (f1) [above left=of 1,xshift=1.3cm,yshift=-1cm] {\Huge $\mc{F}_1$};
\node (f2) [above left=of 2,xshift=2.3cm,yshift=-1cm] {\Huge $\mc{F}_2$};
\node (f3) [above left=of 3,xshift=2.3cm,yshift=-1cm] {\Huge $\mc{F}_3$};
\node (f4) [above left=of 4,xshift=1.3cm,yshift=-1cm] {\Huge $\mc{F}_4$};
\node (f5) [above left=of 5,xshift=2.3cm,yshift=-1cm] {\Huge $\mc{F}_5$};
\node (f6) [above left=of 13,xshift=1.3cm,yshift=-1cm] {\Huge $\mc{F}_6$};
\node (f7) [above left=of 45,xshift=1.3cm,yshift=-1cm] {\Huge $\mc{F}_7$};
\node (f8) [above left=of 123,xshift=1.3cm,yshift=-1cm] {\Huge $\mc{F}_8$};

\draw[-] (1.north) -- (13.south);
\draw[-] (3.north) -- (13.south);
\draw[-] (13.north) -- (123.south);
\draw[-] (2.north) -- (123.south);
\draw[-] (4.north) -- (45.south);
\draw[-] (5.north) -- (45.south);
\end{tikzpicture}
}
\vspace*{-2mm}
\caption{An example of an LT for 5 problem variables, where each node indicates that a linkage set exists containing the problem variables with the denoted indices.}
\label{fig:linkagetree}
\vspace*{-3mm}
\end{figure}

Various linkage models exist.
The model most commonly adopted, especially in a BBO setting, is the Linkage Tree (LT).
An LT can capture hierarchical dependencies and consists of linkage sets of various sizes, from just a single variable, up to a linkage set of at most $\ell - 1$ variables.
An LT firstly consists of all univariate elements, while all larger linkage sets are the union of exactly two linkage sets.
This hierarchical model is constructed using the hierarchical Unweighted Pair Grouping Method with Arithmetic mean (UPGMA) \cite{gronau2007optimal} clustering algorithm, continuously merging the pairs of linkage sets that are considered to be the most strongly dependent.
Dependence between sets is defined as the pairwise average notion of dependence between all variables in one set and all variables in the other set.
Formally stated, for each linkage set $\mc{F}_i \in \vec{\mc{F}}$ of size larger than 1, there exists exactly one pair of linkage sets $\mc{F}_j,\mc{F}_k \in \vec{\mc{F}}$ such that $\mc{F}_j \cap \mc{F}_k = \emptyset$ and $\mc{F}_j \cup \mc{F}_k = \mc{F}_i$.
An example of an LT is shown in Figure \ref{fig:linkagetree}.
Note that the linkage set containing all problem variables is not included in the LT, because using this set as a crossover mask would simply result in the copying of the donor solution.

An LT is generally learned at the start of each generation based on mutual information in the population \cite{thierens2011optimal}, though it is possible to learn an LT prior to the optimization process based on domain knowledge.
Such an LT is generally referred to as a Fixed Linkage Tree (FLT).
In particular, this is interesting in a GBO setting that allows for partial evaluations, because it is known from the VIG which variables are (in)dependent.
Furthermore, an LT may be bounded, in which case, merges of linkage sets that would create linkage sets above the maximum linkage set size would be avoided, and the learning process halts when no more linkage sets with an allowed size can be created.
In this case, the LT is referred to as a Bounded Fixed Linkage Tree (BFLT).

\subsection{Gene-pool Optimal Mixing}
The key strength of GOMEA comes from its use of the GOM variation operator, with which variation is applied to subsets of variables at a time, determined by the linkage model, and variation steps that lead to a deterioration of the fitness of the parent are rejected
During each generation of GOMEA, the GOM operator is applied to each individual $\vec{x}$ in the population $\mc{P}$, using each linkage $\mc{F}_i$ set in the linkage model $\mc{F}$.
Before applying GOM to the parent solution $\vec{x}$, a donor solution is randomly selected from the population.
A requirement for this donor is that its genotype, restricted to the variables included in the linkage set $\mc{F}_i$, is not identical to that of the parent solution $\vec{x}$.
If no individual in the population adheres to this requirement, this iteration of GOM, i.e., with specified linkage set $\mc{F}_i$ for the specified parent $\vec{x}$, is not applied during this generation.
When GOM is applied to the parent solution $\vec{x}$ using linkage set $\mc{F}_i$ and donor solution $\vec{d}$, all genes specified by the linkage set $\mc{F}_i$ are copied from the donor $\vec{d}$ to the parent $\vec{x}$, and the modification is then evaluated, if possible using a partial evaluation.
Pseudo-code of GOM is displayed in Algorithm \ref{alg:gom}.

\section{Parallel Gene-pool Optimal Mixing}
\subsection{Identifying Parallelization Potential}
When GOM is applied to a specified subset of variables $\mc{F}_i$, a partial evaluation is required that depends only on a subset of variables.
In particular, it depends on all variables in $\mc{F}_i$, and any other variable for which a connection exists in the VIG to a variable in $\mc{F}_i$.
It does not depend on any other variables.
Therefore, a number of such partial evaluations may be performed in parallel.
When the mapping function $g$ in Equation \ref{eq:gbo} is the identity function, the decision whether each of these variation steps needs to be accepted, is also independent, allowing them to be performed in parallel as well.
Moreover, this means that two applications of GOM that do not share any dependent variables, are completely independent, and can therefore be performed in parallel.
For large-scale problems with relatively sparse VIGs, it is possible that \emph{many} such applications of GOM are mutually independent, and may therefore be performed in parallel in the procedure that we name parallel GOM.
Moreover, since partial evaluations within different individuals in the population are also independent, there are potentially many parallel steps possible, which fits well with the computing architecture of GPUs.

Prior to performing parallel GOM, we divide all linkage sets into $k$ groups of mutually independent linkage sets, where we aim to minimize $k$ in order to maximize the potential for parallelizability.
Since each linkage set in such a group is independent of all other linkage sets within the same group, all iterations of GOM with linkage sets within the same group may be performed in parallel.
For this purpose, we define the set $\mc{G} = \{\mc{G}_1,\mc{G}_2,\dots,\mc{G}_k\}$ with $\mc{G}_j \subseteq \mc{F}$ and $\mc{G}_i \cap \mc{G}_j = \emptyset$ for each $\mc{G}_i,\mc{G}_j \in \mc{G}$.
Furthermore, because all linkage sets are distributed into groups, for each linkage set $\mc{F}_j \in \mc{F}$, there exists exactly one $\mc{G}_i \in \mc{G}$ that contains $\mc{F}_j$.

\subsection{Operationalizing Parallelization Potential}
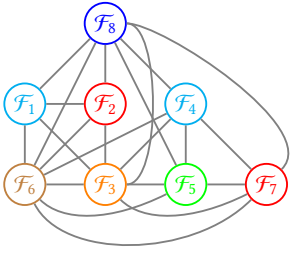
\begin{figure}[h]
\scalebox{0.5}{
\begin{tikzpicture}[transform shape]
    \node[draw,circle,color=cyan,minimum size=1cm,line width=0.5mm] (F1) {\Huge $\mc{F}_1$};
	\node[draw,circle,color=red,minimum size=1cm,line width=0.5mm,right=of F1] (F2) {\Huge $\mc{F}_2$};
    \node[draw,circle,color=orange,minimum size=1cm,line width=0.5mm,below=of F2] (F3) {\Huge $\mc{F}_3$};
    \node[draw,circle,color=cyan,minimum size=1cm,line width=0.5mm,right=of F2] (F4) {\Huge $\mc{F}_4$};
    \node[draw,circle,color=green,minimum size=1cm,line width=0.5mm,below=of F4] (F5) {\Huge $\mc{F}_5$};
    \node[draw,circle,color=brown,minimum size=1cm,line width=0.5mm,below=of F1] (F6) {\Huge $\mc{F}_6$};
    \node[draw,circle,color=red,minimum size=1cm,line width=0.5mm,right=of F5] (F7) {\Huge $\mc{F}_7$};
    \node[draw,circle,color=blue,minimum size=1cm,line width=0.5mm,above=of F2] (F8) {\Huge $\mc{F}_8$};
        
    \draw[gray,line width=0.5mm] (F1) -- (F2);
    \draw[gray,line width=0.5mm] (F1) -- (F3);
	\draw[gray,line width=0.5mm] (F2) -- (F3);
	\draw[gray,line width=0.5mm] (F3) -- (F4);
	\draw[gray,line width=0.5mm] (F3) -- (F5);
	\draw[gray,line width=0.5mm] (F4) -- (F5);
	\draw[gray,line width=0.5mm] (F1) -- (F6);
	\draw[gray,line width=0.5mm] (F3) -- (F6);
	\draw[gray,line width=0.5mm] (F1) -- (F8);
	\draw[gray,line width=0.5mm] (F2) -- (F8);
	\draw[gray,line width=0.5mm] (F2) -- (F6);
	\draw[gray,line width=0.5mm] (F3) .. controls +(1.5,0) and +(1.5,0) .. (F8);
	\draw[gray,line width=0.5mm] (F6) -- (F8);
	\draw[gray,line width=0.5mm] (F4) -- (F7);
	\draw[gray,line width=0.5mm] (F5) -- (F7);
	\draw[gray,line width=0.5mm] (F4) -- (F8);
	\draw[gray,line width=0.5mm] (F5) -- (F8);
	\draw[gray,line width=0.5mm] (F4) -- (F6);
	\draw[gray,line width=0.5mm] (F7) .. controls +(1.5,1.5) and +(1.5,0) .. (F8);
	\draw[gray,line width=0.5mm] (F3) .. controls +(1,-1) and +(-2,-1) .. (F7);
	\draw[gray,line width=0.5mm] (F6) .. controls +(1,-2) and +(-2,-2) .. (F7);
	\draw[gray,line width=0.5mm] (F6) .. controls +(1,-1) and +(-2,-1) .. (F5);
\end{tikzpicture}
}
\vspace*{-3mm}
\caption{The LMIG of a 5-dimensional problem, given the VIG displayed in Figure \ref{fig:vig}, and the linkage model displayed in Figure \ref{fig:linkagetree}. A possible graph coloring is shown that defines the distribution of linkage sets into the groups $\mc{G} = \{\{\mc{F}_1,\mc{F}_4\},\{\mc{F}_2,\mc{F}_7\}, \{\mc{F}_3\}, \{\mc{F}_5\}, \{\mc{F}_6\}, \{\mc{F}_8\} \}$. Note that there is a limited number of linkage sets with the same color, because this example is very small.}
\label{fig:lmig}
\vspace*{-3mm}
\end{figure}

In order to divide the linkage model into the smallest possible number of mutually independent sets, we first create a graph $\mc{L}$ similar to the VIG of the optimization problem.
In contrast to the VIG, the graph $\mc{L}$ shows interactions between linkage sets instead of interactions between variables.
We therefore name this graph the Linkage Model Interaction Graph (LMIG).
We define the graph $\mc{L} = (\vec{V},\vec{E})$ with $\vec{V} = [1,2,\dots,|\mc{F}|]$ and $\vec{E}$ such that there exists an edge $(i,j) \in \vec{E}$ iff linkage sets $\mc{F}_i$ and $\mc{F}_j$ are dependent.
Two linkage sets $\mc{F}_i$ and $\mc{F}_j$ are dependent when $\mc{F}_i \cap \mc{F}_j \neq \emptyset$, or an edge $(u,v)$ exists in the VIG such that $u \in \mc{F}_i$ and $v \in \mc{F}_j$.
The problem of finding the minimum number of mutually independent sets within the linkage model $\mc{F}$ is now equivalent to finding the minimum graph coloring of $\mc{L}$.
The application of graph coloring for this purpose within GOMEA was previously suggested in \cite{bouter2021accelerated}, and was, in a similar way, applied to a parallel hill climber for the optimization of NK-landscapes \cite{derbel2021graph}.
Because graph coloring is an NP-complete problem, finding the optimal minimum graph coloring would be computationally infeasible.
Therefore, we use the greedy Welsh-Powell algorithm \cite{welsh1967upper} to find a graph coloring.
An example of a colored LMIG, given the VIG displayed in Figure \ref{fig:vig} and the linkage model displayed in Figure \ref{fig:linkagetree}, is shown in Figure \ref{fig:lmig}.

\subsection{GPU Implementation}
Pseudo-code of parallel GOMEA is displayed in Algorithm \ref{alg:pargomea}.
In the function $\texttt{DetermineAndInsertDonorGenes}$, firstly, a suitable donor from the population $\mc{P}$ is selected for each of the $n|\mc{G}_i|$ iterations of GOM, and inserted into a copy of the offspring $\mc{O}'$.
The selection of the donor for a specific GOM iteration, i.e., for a specific individual and a specific linkage set, is done by one thread block, because this enables the use of all threads within the thread block to check whether the donor is equal to the parent, which is required by GOM.
Secondly, once a donor has been found that is not equal to the parent, a copy of its genes, restricted to those in the specified linkage set, are inserted into the parent within a copy of the offspring population, named $\mc{O}'$.
The partial fitness contributions for each of the $n|\mc{G}_i|$ modifications caused by $\texttt{DetermineAndInsertDonorGenes}$ are then evaluated in parallel in the $\texttt{ParallelPartialEvaluations}$ function.
For each of the modifications, the fitness of the parent is updated, and the fitness contribution is stored in the $n \times |\mc{G}_i|$ matrix $\vec{\Delta f_{\vec{O}}}$, which stores the contribution of the GOM iteration with linkage set $\mc{F}_j$ applied to individual $\mc{P}[i]$ at position $[i,j]$.
This matrix is then used to determine which of the variation steps must be accepted, which is done in the function $\texttt{DetermineImprovements}$.
In this function, the $n \times |\mc{G}_i|$ matrix $\vec{M}^{\texttt{IMP}}$ is computed, which contains a 1 at position $[i,j]$ when the GOM iteration linkage set $\mc{F}_j$ applied to individual $\mc{P}[i]$ must be accepted.
This is the case when $\vec{\Delta f_{\vec{O}'}}[i,j]$ is larger than 0, or when $\vec{\Delta f_{\vec{O}'}}[i,j]$ is equal to 0 and the individual $\mc{P}[i]$ is not equal to the elitist solution.
Otherwise, the GOM iteration must be rejected, and a 0 is placed in position $[i,j]$ of the matrix $\vec{M}^{\texttt{IMP}}$.
Since all variation steps, those that are accepted and those that are rejected, have been applied to $\mc{O}'$ in the function $\texttt{DetermineAndInsertDonorGenes}$, the rejected variation steps now need to be restored to their backup state $\mc{O}$, which is done in the function $\texttt{ResetNonImprovements}$.
Similarly, in the function $\texttt{InsertImprovements}$, all accepted variation steps are applied to the population, meaning that the offspring $\mc{O}$ and its copy $\mc{O}'$ are now exact copies again.

In particular, pseudocode for the $\texttt{ParallelPartialEvaluations}$ is displayed in Algorithm \ref{alg:parevals}.
This procedure starts with finding all subfunctions dependent on the linkage sets to which GOM is applied in parallel, and the initialization of a matrix of keys for each of these sets subfunctions, such that each subfunction has a key that uniquely depends on the dependent linkage set and the parent solution.
All subfunctions in the list of dependent subfunctions are then evaluated in parallel for all solutions in the offspring, i.e., the solutions to which variation has been applied and are required to be evaluated, and their results stored in a matrix of fitness value contributions.
In parallel, for all solutions in the population, the values of all dependent subfunctions are evaluated.
These results are then subtracted from the matrix of fitness value contributions, meaning that this matrix now describes, for each subfunction, the difference in fitness contribution caused by the respective variation step.
Note that it is possible to store the fitness contribution values of all subfunctions of the population to avoid their recalculation, though this requires a larger amount of memory.

The thrust library \cite{bell2011thrust} is then used to apply a parallel reduction by key, using the keys matrix and the fitness value contribution matrix.
This results in a pair of vectors, one with all occurring keys, and one with the total sum of all fitness contributions for each of the occurring keys.
These vectors are transformed to the matrix $\vec{\Delta f_{\vec{O}}}$ such that each element contains the change in fitness caused the one variation step of GOM applied to a particular solution in the population.

\begin{small}
\begin{algorithm}
\caption{Parallel GOMEA (GBO)}\label{alg:pargomea}
\begin{algorithmic}[1]
\Procedure{$\texttt{Par-GOMEA}$}{$n,\texttt{VIG}$}
\State $\mc{P} \gets \texttt{InitializePopulation}(n)$
\State $\mc{F} \gets \texttt{LearnLinkageModel}(\texttt{VIG})$
\State $\mc{G} \gets \texttt{GraphColoring}(\mc{F})$
\While { $\textbf{not}\;\texttt{TerminationCriterionSatisfied}()$ }
  \State $\mc{O} \gets \texttt{Clone}(\mc{P})$
  \For {$\mc{G}_i \in \mc{G}$} \Comment{Random order}
  \State $\mc{O}' \gets \texttt{DetermineAndInsertDonorGenes}(\mc{G}_i,\mc{O},\mc{P})$
  \State $\vec{\Delta f_{\mc{O}'}} \gets \texttt{ParallelPartialEvaluations}(\mc{G}_i,\mc{O},\mc{O}')$
  \State $\vec{M}^{\texttt{IMP}} \gets \texttt{DetermineImprovements}(\mc{G}_i,\mc{O}',\Delta f_{\mc{O}'})$
  \State $\mc{O}' \gets \texttt{ResetNonImprovements}(\mc{G}_i,\mc{O},\vec{M}^{\texttt{IMP}})$
  \State $\mc{O} \gets \texttt{InsertImprovements}(\mc{G}_i,\mc{O}',\vec{M}^{\texttt{IMP}})$
  \EndFor
  \State $\mc{P} \gets \mc{O}$
\EndWhile
\EndProcedure
\end{algorithmic}
\end{algorithm}
\end{small}

\begin{small}
\begin{algorithm}
\caption{Parallel Partial Evaluations}\label{alg:parevals}
\begin{algorithmic}[1]
\Procedure{$\texttt{ParallelPartialEvaluations}$}{$\mc{G}_i,\mc{O},\mc{O}'$}
\State $\vec{F}^\texttt{dep}, \vec{K}^\texttt{dep},\vec{M}^\texttt{dep}, \vec{\Delta f_{\vec{O}}} \gets []$
\State // Determine subfunctions to be evaluated
\For{$\mc{F}_j \in \mc{G}_i$}
	\State $\vec{F}_j \gets \emptyset$ \Comment{Dependent subfunctions of $\mc{F}_j$}
	\For{$u \in \mc{F}_j$}
		\State $\vec{F}_j \gets \vec{F}_j \cup \texttt{DependentSubfunctions}(u)$
	\EndFor
	\For{$u \in \vec{F}_j$}
		\State $\vec{F}^\texttt{dep}.\texttt{append}(u)$ \Comment{Dependent subfunctions}
		\State $\vec{K}^\texttt{dep}.\texttt{append}(j)$ \Comment{Respective linkage sets}
	\EndFor
\EndFor

\State // Create matrix of unique keys
\ParFor{$u \in [1,2,\dots,|\vec{K}^\texttt{dep}|]$}
	\ParFor{$v \in [1,2,\dots,|\mc{O}'|]$}
		\State $\vec{M}^\texttt{dep}[u,v] \gets \vec{K}^\texttt{dep}[u] + v \cdot |\vec{K}^\texttt{dep}|$
	\EndParFor
\EndParFor

\State // Evaluate dependent subfunctions
\ParFor{$u \in [1,2,\dots,|\mc{O}|]$} \Comment{$\mc{O}$ and $\mc{O}'$ have equal sizes}
	\ParFor{$v \in \vec{F}^\texttt{dep}$} \Comment{Dependent subfunctions}
		\State $\vec{F}_{\mc{O}}^{part}[u,v] \gets \texttt{EvaluateSubfunction}(\mc{O}_u,v)$
		\State $\vec{F}_{\mc{O}'}^{part}[u,v] \gets \texttt{EvaluateSubfunction}(\mc{O}'_u,v)$
	\EndParFor
\EndParFor

\State // Determine sum of evaluated subfunctions
\State $\vec{F}_{\mc{O}}^\texttt{out} \gets \texttt{ReduceByKey}(\vec{F}_{\mc{O}}^{part},\vec{M}^\texttt{dep})$
\State $\vec{F}_{\mc{O}'}^\texttt{out} \gets \texttt{ReduceByKey}(\vec{F}_{\mc{O}'}^{part},\vec{M}^\texttt{dep})$

\State // Matrix of fitness change for all variation steps
\ParFor{$u \in [1,2,\dots,|\vec{K}^\texttt{dep}|]$}
	\ParFor{$v \in [1,2,\dots,|\mc{O}'|]$}
		\State $\texttt{ind}_f \gets \vec{K}^\texttt{dep}[u] + v \cdot |\vec{K}^\texttt{dep}|$
		\State $\vec{\Delta f_{\vec{O}}}[u,v] \gets \vec{F}_{\mc{O}'}^\texttt{out}[\texttt{ind}_f] - \vec{F}_{\mc{O}}^\texttt{out}[\texttt{ind}_f]$
	\EndParFor
\EndParFor
	
\State \Return{$\vec{\Delta f_{\vec{O}}}[u,v]$}
\EndProcedure
\end{algorithmic}
\end{algorithm}
\end{small}

\subsection{Differences with Serial GOM}
\label{subsec:parserdiff}
Due to the serial nature of GOM, and the SIMD architecture of GPUs, some aspects of GOMEA do not translate well to GPUs.
Therefore, some of these aspects of GOMEA have been adapted or left out, because they would have a large impact on the parallelization potential of GOMEA when left unchanged.

Firstly, in the serial GOMEA, GOM is applied to one individual with all linkage sets, before moving to the next individual.
Because this order of operations would leave the parallelization potential of the population unused, this order is changed in the parallel GOM, opening the possibility that a large number of steps of GOM are performed for each of the solutions in the population in parallel.
Additionally, because linkage sets are divided into mutually independent groups, the order in which GOM is applied to the population is no longer uniformly at random.
Instead, only the order of the groups is uniformly at random, which can potentially introduce a bias.
These changes in the order of GOM are, however, unavoidable to enable parallelization.

Secondly, learning an LT based on the population at the start of each generation is computationally expensive, easily becoming a bottleneck when the amount of computation time dedicated to GOM is largely reduced.
Instead, a fixed LT can be learned prior to optimization based on domain knowledge, greatly reducing required computation time.
It is, however, possible to parallelize the UPGMA procedure that is used for the construction of the LT~\cite{chen2012parallel}, which we consider a potential topic of future work.

Thirdly, the Forced Improvement (FI) procedure is disabled in the parallel version of GOMEA, because it is by nature a serial procedure that halts as soon as one of the steps results in an improvement.
Furthermore, FI is generally applied to only a small number of solutions at once, unlike GOM.
We consider the design of a procedure akin to FI that is more amenable to parallelization also a topic of future work.

Any of these changes may have a potential effect on the convergence of GOMEA.
Therefore, they are evaluated in Section \ref{subsec:exp-parserdiff}.

\section{Experiments}
\label{sec:experiments}
In this section, we benchmark the performance of parallel GOM, and compare it to that of the original, serial, GOMEA.
The set-up of these experiments is first described in Section \ref{subsec:setup}, followed by the description of the benchmark problems in Section \ref{subsec:benchmarks}.
We then described various experiments in Sections \ref{subsec:ims} through \ref{subsec:speedup}.

\subsection{Overall Set-up}
\label{subsec:setup}
All experiments with the serial version of GOMEA are executed on a an Intel Xeon CPU E5-2630 v4 core with a clock frequency of 2.20GHz.
The experiments performed on a GPU are executed on an NVIDIA Geforce RTX 2080 Ti, which consists of 4352 CUDA cores at a frequency of 1.54 GHz, and 11 GB of global memory.
For the population size parameter, we use the Interleaved Multi-start Scheme (IMS) (see Section \ref{subsec:ims}).
Unless otherwise mentioned, all versions of GOMEA use an LT learned based on the Max-Cut graph with UPGMA using the weights of the graph as a notion of similarity.
Furthermore, default parameters are used unless specified.
We refer to the CUDA implementation of GOMEA, using parallel GOM, as parallel GOMEA, and the original, serial version of GOMEA, as serial GOMEA.

\subsection{Benchmark Problems}
\label{subsec:benchmarks}
We focus on the well-known NP-complete (Weighted) Maximum Cut (Max-Cut) problem \cite{karp1972reducibility}, because it has a clearly defined dependency structure that allows for a clear demonstration of the use of parallel GOM.
Furthermore, all techniques introduced in this paper directly apply to other optimization problems for which partial evaluations are possible, and, consequently, the VIG is known.

The objective of the Max-Cut problem is, given a weighted graph $G=(V,E)$, to assign each vertex in $V$ to a set $\mc{S}$ or its complement, and to maximize the total sum of weights of the edges between vertices in complementing sets.
Formally, given a graph $G=(V,E)$, the weight function $w(i,j)$ that defines the weight of the edge between each pair of vertices $i$ and $j$ in $V$, and a binary solution $\vec{x}$ of length $\ell=|V|$, the optimization function, subject to maximization, of the Max-Cut problem is defined as follows:

\vspace*{-3mm}
\begin{small}
\begin{align}
\textbf{max}\;\; f^{\texttt{Max-Cut}}(\vec{x}) &= \frac{1}{2} \sum_{i,j \in V;i<j} w_{ij} \left( 1 - x_i x_j \right), &\\
\textbf{s.t.}\;\;w_{ij} & \in \mathbb{R}, & i,j \in V, \\
x_i,x_j & \in [-1,1], & i,j \in V.
\end{align}
\end{small}

Generally speaking, in a GA, each solution $\vec{x}$ has binary variables $x_i \in [0,1]$, but this is trivially mapped to the domain $[-1,1]$.

For our experiments, we use three sets of Max-Cut instances with different properties: Set A, Set B, and Set C.
\begin{itemize}
\item Set A consists of fully connected graphs with 6 up to 200 vertices, and 15 up to 19,900 edges.
\item Set B consists of graphs on a 3D torus, i.e., a 2D grid with wrap-around on all edges, with 9 up to 1600 vertices, and 18 up to 3200 edges.
For each graph in Set B, the connectivity, i.e., number of connected edges, of each vertex is equal to 4.
\item Set C consists of a selection of graphs from the BIQMAC library \cite{rendl2010solving}, named g1, g22, g55, g60, g65, g66, g72, g77, and g81.
These graphs contain 800, 2000, 5000, 7000, 8000, 9000, 10000, 14000, and 20000 vertices, respectively.
A table of all properties is included in the supplementary material.
\end{itemize}

\subsection{Interleaved Multi-start Scheme}
\label{subsec:ims}
To prevent having to tune the population size of each of the algorithms by hand, we use the Interleaved Multi-start Scheme (IMS) \cite{bouter2021achieving}.
In this scheme, the generations of multiple populations of different sizes are performed in an interleaved way, with smaller populations iterating through generations at a higher frequency.
The smallest population in the IMS, denoted $\mc{P}_1$, starts at the base population size $n^{\texttt{base}}$, and each newly initialized population is double the size of the previously largest population, i.e., $|\mc{P}_i| = 2^{i-1} n^{\texttt{base}}$.
Generations of all populations are performed recursively, such that one generation of population $\mc{P}_{i}$ is performed as soon as population $\mc{P}_{i-1}$ has performed $c^{\texttt{IMS}}$ generations.

\begin{figure}[h]
\centering
\begin{subfigure}[t]{0.49\linewidth}
\centering
\includegraphics[width=\linewidth]{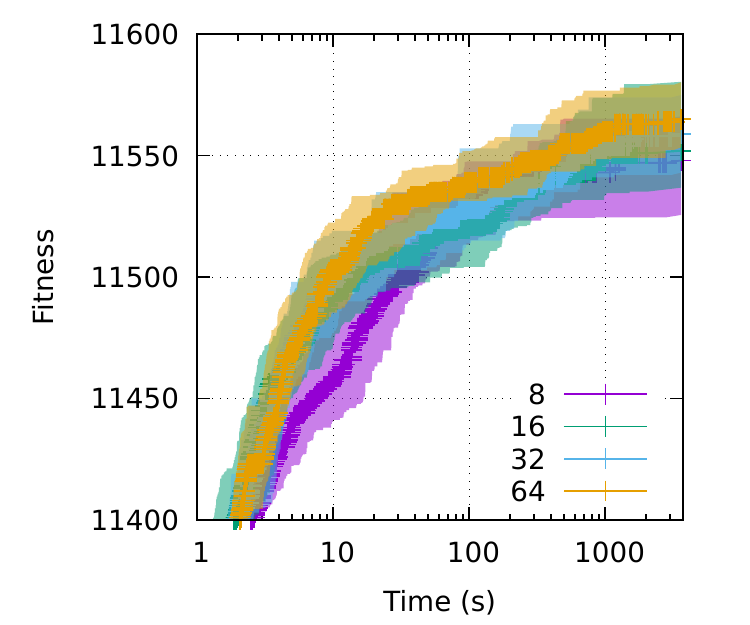}
\label{fig:popsize-g1}
\end{subfigure}
\begin{subfigure}[t]{0.49\linewidth}
\centering
\includegraphics[width=\linewidth]{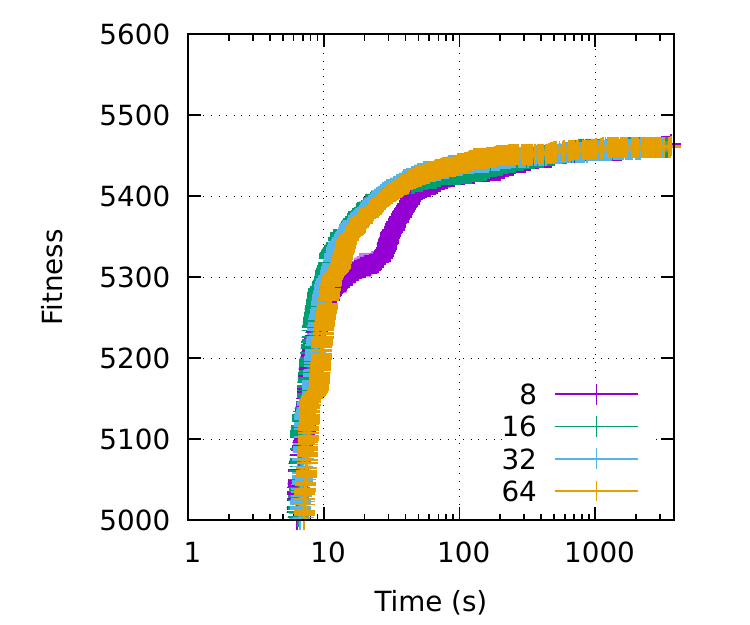}
\label{fig:popsize-g65}
\end{subfigure}
\vspace*{-6mm}
\caption{Median and interdecile range (30 runs) of fitness values achieved of different settings for $n^{\texttt{base}}$ in the IMS for parallel GOMEA using a fixed LT, for g1 and g65 (Set C), respectively.}
\label{fig:ims-popsize}
\vspace*{-3mm}
\end{figure}

Though $n^{\texttt{base}}$ and $c^{\texttt{IMS}}$ are still parameters that may be tuned, their effect on the performance of an EA is much smaller than that of the population size parameter, because a population size that is too small may lead to premature convergence, while this is not the case within the IMS.
In this paper, we use the default setting of $c^{\texttt{IMS}} = 4$, and we show the difference in performance of parallel GOMEA for $n^{\texttt{base}}$ in Figure \ref{fig:ims-popsize}.
This figure shows convergence for different population sizes on different instances in set C, using a time limit of an hour.
For each of the settings, a fixed LT was used.
Based on Figure \ref{fig:ims-popsize}, we find that the setting of $n^{\texttt{base}}$ has only a marginal impact on the performance of parallel GOMEA, with only $n^{\texttt{base}} = 8$ giving slightly worse results.
Similar results were found for different instances.
We therefore use $n^{\texttt{base}} = 16$ in the remainder of our experiments.

\subsection{Scalability}
\begin{figure}[h]
\centering
\begin{subfigure}[t]{0.49\linewidth}
\centering
\includegraphics[width=\linewidth]{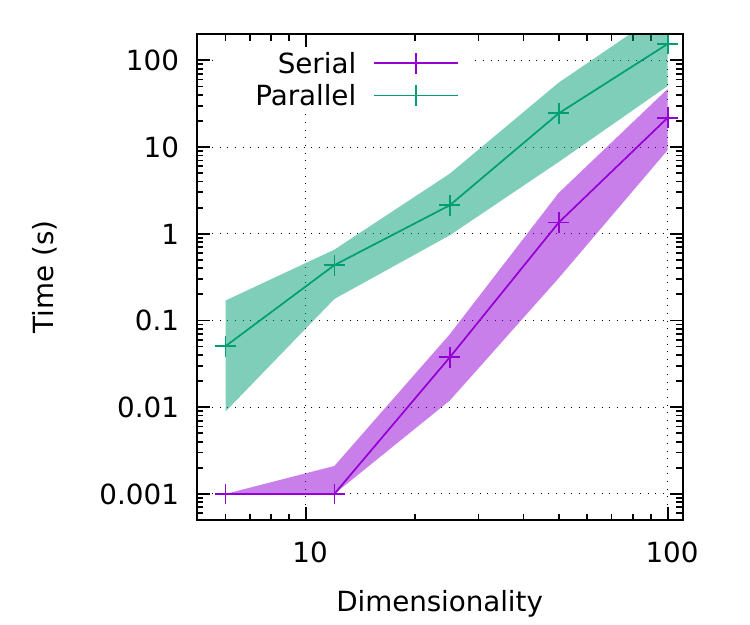}
\caption{Set A}
\label{fig:scalability-setA}
\end{subfigure}
\begin{subfigure}[t]{0.49\linewidth}
\centering
\includegraphics[width=\linewidth]{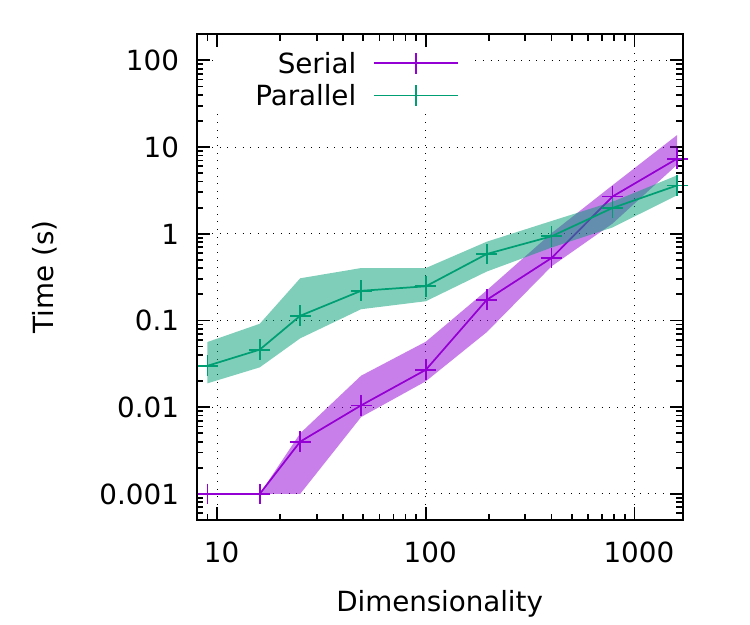}
\caption{Set B}
\label{fig:scalability-setB}
\end{subfigure}
\vspace*{-3mm}
\caption{Scalability of serial and parallel GOMEA, showing the median and interdecile range (30 runs) of the time required to find the optimum, for the instances in set A and set B, respectively.}
\label{fig:scalability}
\vspace*{-3mm}
\end{figure}

In this section, we test the scalability GOMEA with and without the use of parallel GOM.
This is done on instances from Set A and Set B, because these instances have a fixed structure, allowing us to scale up the number of vertices while keeping the structure of the instances constant.
In Figure \ref{fig:scalability} we show what the effect of parallel GOM is on the scalability of GOMEA.

No efficiency improvement was observed for fully connected graphs (see Figure \ref{fig:scalability-setA}), because no iterations of GOM may be performed in parallel.
Though GOM may still be performed in parallel for different individuals in the population, this leaves a large part of the GPU idle, while keeping the overhead from copying memory to and from the GPU device.
Therefore, we find that the GPU implementation of GOMEA performs worse than the serial version of GOMEA on fully connected graphs.
For graphs with a torus-like structure, as displayed in Figure \ref{fig:scalability-setB}, many FOS elements can be subjected to GOM in parallel. Concordantly, we observe a clear difference in scalability. Still, within the scope of available problem sizes, only a small increase in performance is observed over serial GOMEA. No optimum was known for graphs with more than 1600 nodes, prohibiting extending the scalability analysis, although from the results it is to be expected that large speed-ups may well be possible for larger graphs, which we consider in the next subsection.

\subsection{Convergence Speed}
\label{subsec:speedup}
\begin{figure}[h]
\vspace*{-2mm}
\centering
\begin{subfigure}[t]{0.49\linewidth}
\centering
\includegraphics[width=\linewidth]{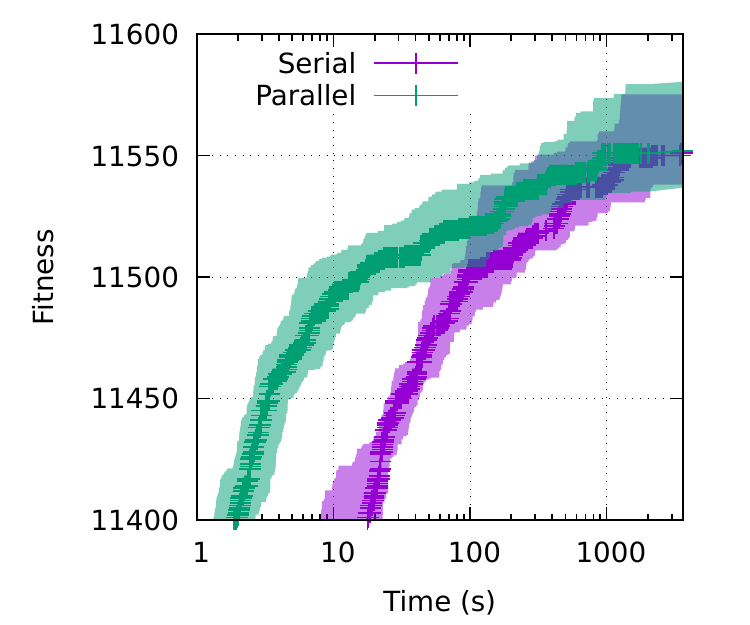}
\label{fig:convergence-g1}
\end{subfigure}
\begin{subfigure}[t]{0.49\linewidth}
\centering
\includegraphics[width=\linewidth]{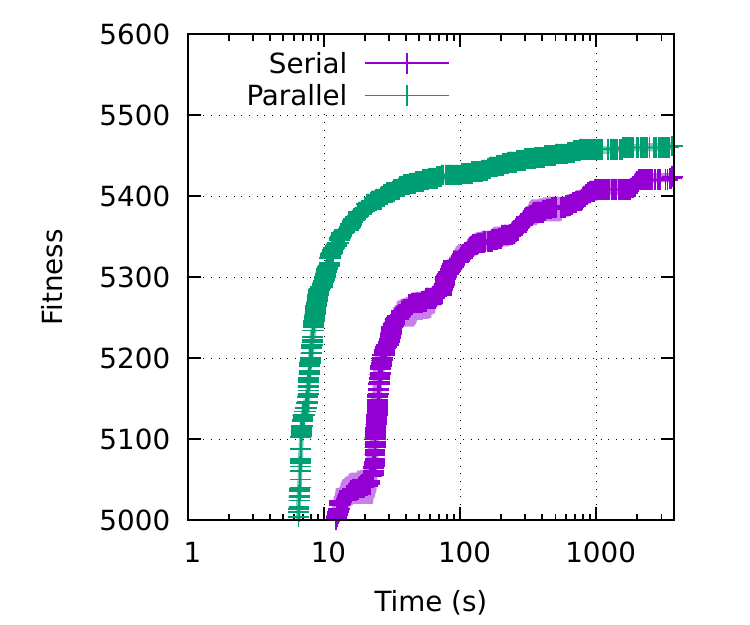}
\label{fig:convergence-g65}
\end{subfigure}
\vspace{-20px}
\vspace*{-1mm}
\caption{Median and interdecile range (30 runs) of fitness values achieved by serial and parallel GOMEA for g1 and g65 (Set C), respectively.}
\label{fig:convergence}
\vspace*{-3mm}
\end{figure}

In this section, we analyze how the use of parallel GOMEA influences the speed of convergence of GOMEA and whether on larger graphs, GOMEA with the best settings within our time budget can still be run faster on a GPU.
Firstly, for 2 instances, we show convergence plots in Figure \ref{fig:convergence}, comparing GOMEA with and without the use of parallel GOM.
Convergence plots for the other instances are provided in the supplementary material.

Secondly, in Figure \ref{fig:speedups}, we show speed-ups achieved by using parallel GOMEA, compared to serial GOMEA.
This Figure is split, showing speed-ups for the five smallest instances on the left, and speed-ups for the four largest instances on the right.
For the smaller instances, a time limit of 1 hour was used for both serial and parallel GOMEA.
For the larger instances, a time limit of 6 hours was used for serial GOMEA, and 1 hour for parallel GOMEA.
On the horizontal axis, we show the (median of 30 runs) amount of time required by parallel GOMEA to reach a certain fitness value.
The vertical axis shows the factor by which serial GOMEA required more time (median of 30 runs) than parallel GOMEA, to achieve the same fitness value, i.e., the parallel GOMEA speed-up factor.

We find that, for large instances, speed-ups up to a factor of 100 can be achieved. Likely, if run even longer and for even larger instances, the speed-up factor could be even larger.
On the five smallest instances, parallel GOMEA generally achieves a speed-up at the start of the optimization process, but this speed-up decreases over time.
This is likely a consequence of the differences between serial and parallel GOMEA, as discussed in Section \ref{subsec:parserdiff}.

\begin{figure}[h]
\centering
\begin{subfigure}[t]{0.49\linewidth}
\centering
\includegraphics[width=\linewidth]{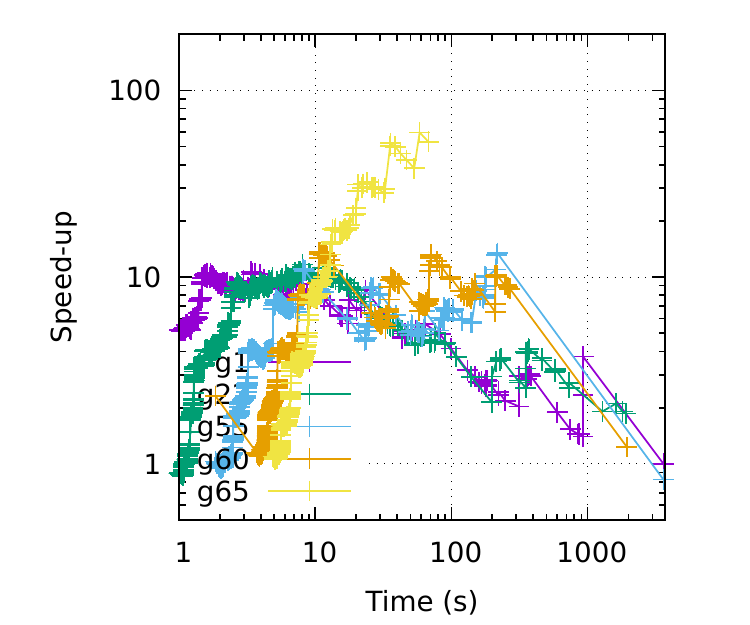}
\label{fig:speedup-small}
\end{subfigure}
\begin{subfigure}[t]{0.49\linewidth}
\centering
\includegraphics[width=\linewidth]{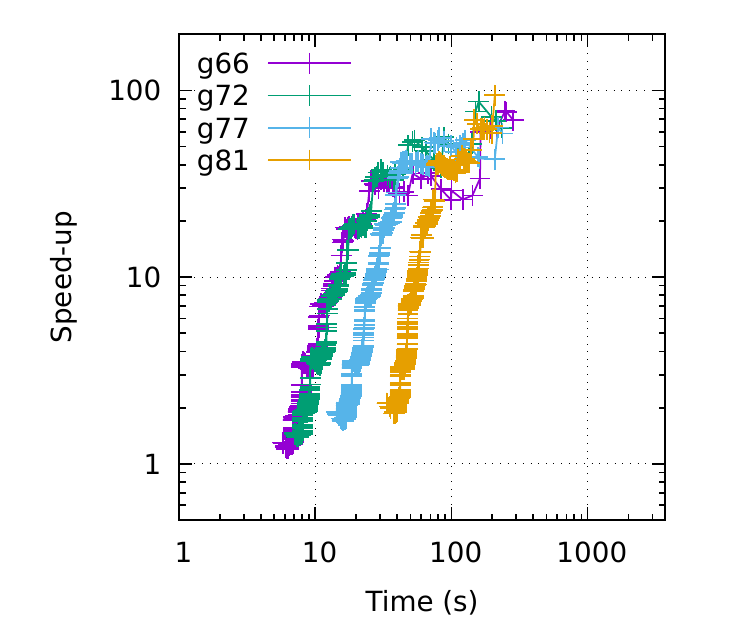}
\label{fig:speedup-large}
\end{subfigure}
\vspace{-20px}
\vspace*{-1mm}
\caption{Median (30 runs) speed-up achieved by parallel GOMEA, compared to serial GOMEA, for set C.}
\label{fig:speedups}
\vspace*{-3mm}
\end{figure}

\subsection{Parallel and Serial GOMEA Differences}
\label{subsec:exp-parserdiff}
In this section, we analyze the difference in convergence for parallel and serial GOMEA from an algorithmic point of view. That is, all experiments in this Section are run with a serial version of GOMEA, but with changes to some of its operations to reflect algorithmic differences with the GPU parallel version of GOMEA.

\begin{figure}[h]
\centering
\begin{subfigure}[t]{0.49\linewidth}
\centering
\includegraphics[width=\linewidth]{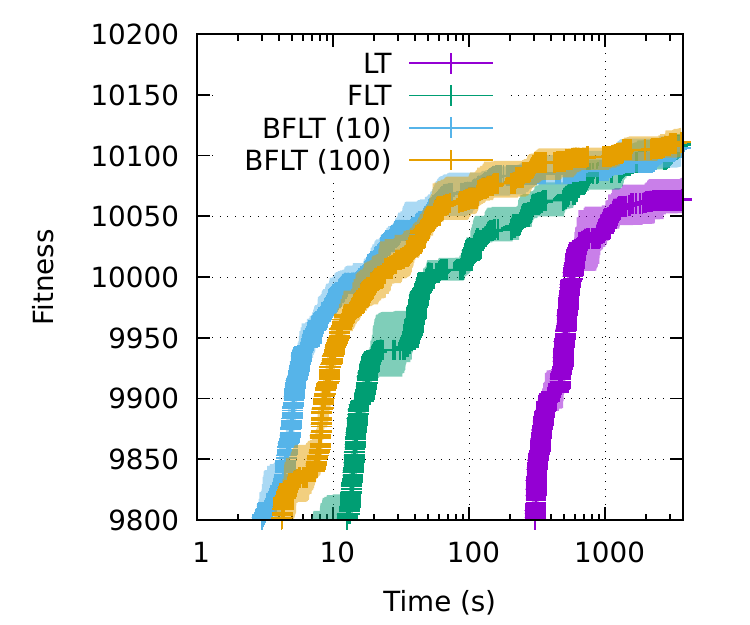}
\label{fig:linkagemodels-g55}
\end{subfigure}
\begin{subfigure}[t]{0.49\linewidth}
\centering
\includegraphics[width=\linewidth]{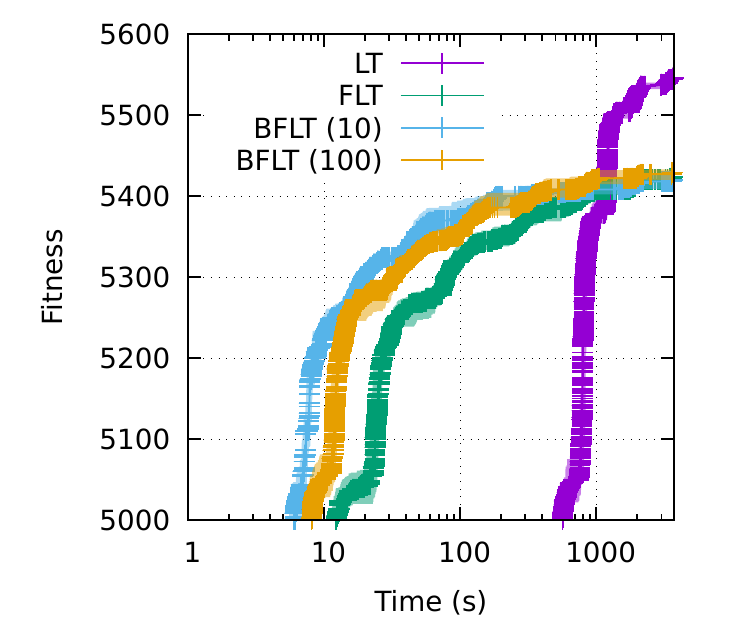}
\label{fig:linkagemodels-g65}
\end{subfigure}
\vspace{-20px}
\caption{Median and interdecile range (30 runs) of fitness values achieved by serial GOMEA using different linkage models for g55 and g65 (Set C), respectively.}
\label{fig:linkagemodels}
\end{figure}

Firstly, we test the impact of using different linkage models.
This includes a Linkage Tree (LT) learned from the population using mutual information at the start of each generation, an FLT learned using UPGMA with weights of the Max-Cut graph as a similarity measure, and BFLTs bounded by 10 and 100. Convergence results pertaining to different linkage models are shown in Figure \ref{fig:linkagemodels}.
This Figure shows that, in particular for instance g65, using a static linkage tree may lead to premature convergence, as using the linkage tree that is learned at the start of every generation clearly performs better, which is in-line with existing literature that showed better performance using learned LTs versus fixed LTs on linkage benchmark problems~\cite{thierens2012predetermined}. However, better results are obtained  much later in terms of time due to the need to, every generation, estimate a large mutual information matrix and, based on this, create an LT.
Results on instance g55 show that the added value of learning an LT during search may come even later in the search process, as within our time limit it led to the worst results.

\begin{figure}[h]
\centering
\begin{subfigure}[t]{0.49\linewidth}
\centering
\includegraphics[width=\linewidth]{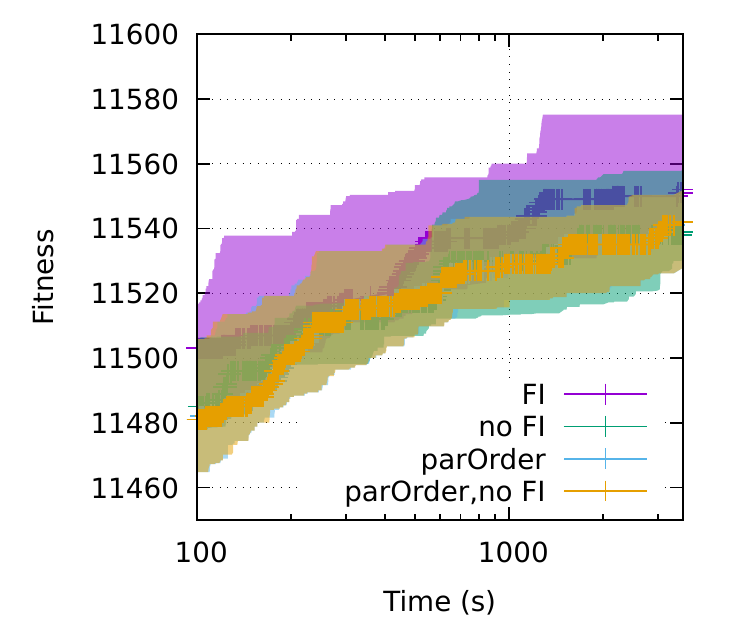}
\label{fig:fosorder-g1}
\end{subfigure}
\begin{subfigure}[t]{0.49\linewidth}
\centering
\includegraphics[width=\linewidth]{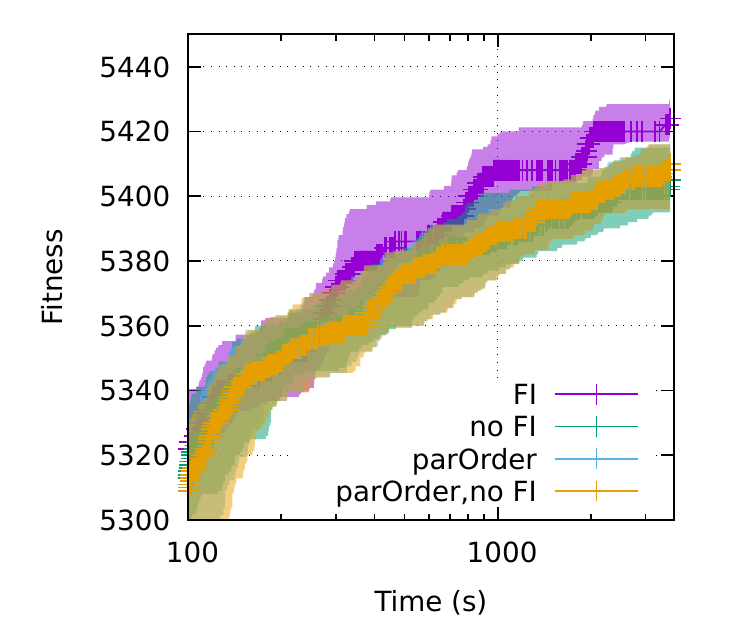}
\label{fig:fosorder-g65}
\end{subfigure}
\vspace{-20px}
\caption{Median and interdecile range (30 runs) of fitness values achieved by serial GOMEA using a random FOS order, or one dictated by parallel GOM (parOrder), and with FI enabled or disabled, for g1 and g65 (Set C), respectively.}
\label{fig:fosorder}
\end{figure}

Secondly, in Figure \ref{fig:fosorder}, we show how the order of GOM and the FI procedure, which are different in the parallel GOMEA as discussed in Section \ref{subsec:parserdiff}, influence performance.
Note that these figures are zoomed in to the later stages of convergence (after 100 seconds).
We observe that, while the graph-coloring-based group-wise order of processing the FOS needed to realize large-scale parallelization does not appear to have an impact on performance of GOMEA, the FI procedure does, allowing it to converge to better fitness values in the very late stages of the optimization process. This is also the reason why the speed-ups obtained by parallel GOMEA as observed in Figure \ref{fig:speedups}, disappear at a later stage of the search process. Even though the parallel GOMEA then still performs many more evaluations per second than the serial GOMEA, this benefits is outweighed by the algorithmic added value of FI.

\section{Discussion and conclusion}
The main purpose of this paper was to show the potential of parallel GOM, for which the Max-Cut problem was selected as a benchmark, because it is a well-known problem with a clear structure that enables a GBO setting.
Even so, we note that it is likely that parallel GOM is outperformed by various other methods that are considered state-of-the-art for the Max-Cut problem.
Furthermore, though it is not within the scope of this paper, the performance of GOMEA on the Max-Cut problem may be greatly improved by the addition of (Iterated) Local Search ((I)LS).
Similar to how graph coloring is applied to find independent sets for the application of parallel GOMEA, this can be done to parallelize (I)LS to create a hybrid parallel GOMEA, which is an interesting direction for future work.

In this paper, we showed how the GOM variation operator of the state-of-the-art model-based EA known as GOMEA may be applied in a large-scale parallel manner to apply variation steps to a large number of non-trivial-sized subsets of problem variables for each individual in the population.
Using a CUDA implementation of the so-constructed parallel GOMEA on a GPU, we were able to speed up the performance of GOMEA on the well-known MaxCut problem up to a factor of 100. Likely, larger speed-up factors are possible if even larger graph instances would be used. Moreover, we identified additional potential improvements. Altogether, this paper contributes to the body of empirical evidence that shows that the use of GPUs holds vast potential to accelerate modern, powerful EAs on contemporary computing hardware and have an important impact to the field of EC in general, similar to how they propelled the field of deep learning to new heights \cite{mittal2019survey}.


\bibliographystyle{ACM-Reference-Format}
\bibliography{maxcutgpu}


\begin{thebibliography}{30}


\ifx \showCODEN    \undefined \def \showCODEN     #1{\unskip}     \fi
\ifx \showDOI      \undefined \def \showDOI       #1{#1}\fi
\ifx \showISBNx    \undefined \def \showISBNx     #1{\unskip}     \fi
\ifx \showISBNxiii \undefined \def \showISBNxiii  #1{\unskip}     \fi
\ifx \showISSN     \undefined \def \showISSN      #1{\unskip}     \fi
\ifx \showLCCN     \undefined \def \showLCCN      #1{\unskip}     \fi
\ifx \shownote     \undefined \def \shownote      #1{#1}          \fi
\ifx \showarticletitle \undefined \def \showarticletitle #1{#1}   \fi
\ifx \showURL      \undefined \def \showURL       {\relax}        \fi
\providecommand\bibfield[2]{#2}
\providecommand\bibinfo[2]{#2}
\providecommand\natexlab[1]{#1}
\providecommand\showeprint[2][]{arXiv:#2}

\bibitem[\protect\citeauthoryear{Alba}{Alba}{2006}]%
        {alba2006parallel}
\bibfield{author}{\bibinfo{person}{Enrique Alba}.}
  \bibinfo{year}{2006}\natexlab{}.
\newblock \bibinfo{booktitle}{\emph{Parallel evolutionary computations}}.
  Vol.~\bibinfo{volume}{22}.
\newblock \bibinfo{publisher}{springer}.
\newblock


\bibitem[\protect\citeauthoryear{Asanovic, Bodik, Catanzaro, Gebis, Husbands,
  Keutzer, Patterson, Plishker, Shalf, Williams, et~al\mbox{.}}{Asanovic
  et~al\mbox{.}}{2006}]%
        {asanovic2006landscape}
\bibfield{author}{\bibinfo{person}{Krste Asanovic}, \bibinfo{person}{Ras
  Bodik}, \bibinfo{person}{Bryan~Christopher Catanzaro},
  \bibinfo{person}{Joseph~James Gebis}, \bibinfo{person}{Parry Husbands},
  \bibinfo{person}{Kurt Keutzer}, \bibinfo{person}{David~A Patterson},
  \bibinfo{person}{William~Lester Plishker}, \bibinfo{person}{John Shalf},
  \bibinfo{person}{Samuel~Webb Williams}, {et~al\mbox{.}}}
  \bibinfo{year}{2006}\natexlab{}.
\newblock \showarticletitle{The landscape of parallel computing research: {A}
  view from {Berkeley}}.
\newblock  (\bibinfo{year}{2006}).
\newblock


\bibitem[\protect\citeauthoryear{Bell and Hoberock}{Bell and Hoberock}{2011}]%
        {bell2011thrust}
\bibfield{author}{\bibinfo{person}{N. Bell} {and} \bibinfo{person}{J.
  Hoberock}.} \bibinfo{year}{2011}\natexlab{}.
\newblock \showarticletitle{Thrust: {A} productivity-oriented library for
  {CUDA}}.
\newblock \bibinfo{journal}{\emph{{GPU} computing gems {Jade} edition}}
  \bibinfo{volume}{2} (\bibinfo{year}{2011}), \bibinfo{pages}{359--371}.
\newblock


\bibitem[\protect\citeauthoryear{Bosman and Thierens}{Bosman and
  Thierens}{2012}]%
        {bosman2012linkage}
\bibfield{author}{\bibinfo{person}{P.~A.~N. Bosman} {and} \bibinfo{person}{D.
  Thierens}.} \bibinfo{year}{2012}\natexlab{}.
\newblock \showarticletitle{Linkage neighbors, optimal mixing and forced
  improvements in genetic algorithms}. In \bibinfo{booktitle}{\emph{Proc.
  GECCO}}. ACM, \bibinfo{pages}{585--592}.
\newblock


\bibitem[\protect\citeauthoryear{{Bouter, A.}, Alderliesten, Bel, Witteveen,
  and Bosman}{{Bouter, A.} et~al\mbox{.}}{2018}]%
        {bouter2018large}
\bibfield{author}{\bibinfo{person}{{Bouter, A.}}, \bibinfo{person}{Tanja
  Alderliesten}, \bibinfo{person}{Arjan Bel}, \bibinfo{person}{Cees Witteveen},
  {and} \bibinfo{person}{Peter A~N Bosman}.} \bibinfo{year}{2018}\natexlab{}.
\newblock \showarticletitle{Large-scale parallelization of partial evaluations
  in evolutionary algorithms for real-world problems}. In
  \bibinfo{booktitle}{\emph{Proc. GECCO}}. ACM, \bibinfo{pages}{1199--1206}.
\newblock


\bibitem[\protect\citeauthoryear{{Bouter, A.}, Alderliesten, and
  Bosman}{{Bouter, A.} et~al\mbox{.}}{2021a}]%
        {bouter2021achieving}
\bibfield{author}{\bibinfo{person}{{Bouter, A.}}, \bibinfo{person}{Tanja
  Alderliesten}, {and} \bibinfo{person}{Peter A.~N. Bosman}.}
  \bibinfo{year}{2021}\natexlab{a}.
\newblock \showarticletitle{Achieving highly scalable evolutionary real-valued
  optimization by exploiting partial evaluations}.
\newblock \bibinfo{journal}{\emph{Evolutionary computation}}
  \bibinfo{volume}{29}, \bibinfo{number}{1} (\bibinfo{year}{2021}),
  \bibinfo{pages}{129--155}.
\newblock


\bibitem[\protect\citeauthoryear{{Bouter, A.}, Alderliesten, and
  Bosman}{{Bouter, A.} et~al\mbox{.}}{2021b}]%
        {bouter2021accelerated}
\bibfield{author}{\bibinfo{person}{{Bouter, A.}}, \bibinfo{person}{Tanja
  Alderliesten}, {and} \bibinfo{person}{Peter A~N Bosman}.}
  \bibinfo{year}{2021}\natexlab{b}.
\newblock \showarticletitle{{GPU}-Accelerated Parallel Gene-pool Optimal Mixing
  applied to Multi-Objective Deformable Image Registration}. In
  \bibinfo{booktitle}{\emph{2021 IEEE Congress on Evolutionary Computation
  (CEC)}}. IEEE, \bibinfo{pages}{2539--2548}.
\newblock


\bibitem[\protect\citeauthoryear{Brodtkorb, Dyken, Hagen, Hjelmervik, and
  Storaasli}{Brodtkorb et~al\mbox{.}}{2010}]%
        {brodtkorb2010state}
\bibfield{author}{\bibinfo{person}{Andre~R Brodtkorb},
  \bibinfo{person}{Christopher Dyken}, \bibinfo{person}{Trond~R Hagen},
  \bibinfo{person}{Jon~M Hjelmervik}, {and} \bibinfo{person}{Olaf~O
  Storaasli}.} \bibinfo{year}{2010}\natexlab{}.
\newblock \showarticletitle{State-of-the-art in heterogeneous computing}.
\newblock \bibinfo{journal}{\emph{Scientific Programming}}
  \bibinfo{volume}{18}, \bibinfo{number}{1} (\bibinfo{year}{2010}),
  \bibinfo{pages}{1--33}.
\newblock


\bibitem[\protect\citeauthoryear{Cabrera, Ehrgott, Mason, and Philpott}{Cabrera
  et~al\mbox{.}}{2014}]%
        {cabrera2014multi}
\bibfield{author}{\bibinfo{person}{Guillermo Cabrera},
  \bibinfo{person}{Matthias Ehrgott}, \bibinfo{person}{Andrew Mason}, {and}
  \bibinfo{person}{Andy Philpott}.} \bibinfo{year}{2014}\natexlab{}.
\newblock \showarticletitle{Multi-objective optimisation of positively
  homogeneous functions and an application in radiation therapy}.
\newblock \bibinfo{journal}{\emph{Operations Research Letters}}
  \bibinfo{volume}{42}, \bibinfo{number}{4} (\bibinfo{year}{2014}),
  \bibinfo{pages}{268--272}.
\newblock


\bibitem[\protect\citeauthoryear{Chen, Hung, Lin, Lin, Lee, and Lee}{Chen
  et~al\mbox{.}}{2012}]%
        {chen2012parallel}
\bibfield{author}{\bibinfo{person}{Yu-Rong Chen}, \bibinfo{person}{Che~Lun
  Hung}, \bibinfo{person}{Yu-Shiang Lin}, \bibinfo{person}{Chun-Yuan Lin},
  \bibinfo{person}{Tien-Lin Lee}, {and} \bibinfo{person}{Kual-Zheng Lee}.}
  \bibinfo{year}{2012}\natexlab{}.
\newblock \showarticletitle{Parallel UPGMA algorithm on graphics processing
  units using CUDA}. In \bibinfo{booktitle}{\emph{2012 IEEE 14th International
  Conference on High Performance Computing and Communication \& 2012 IEEE 9th
  International Conference on Embedded Software and Systems}}. IEEE,
  \bibinfo{pages}{849--854}.
\newblock


\bibitem[\protect\citeauthoryear{Chicano, Whitley, Ochoa, and
  Tin{\'o}s}{Chicano et~al\mbox{.}}{2017}]%
        {chicano2017optimizing}
\bibfield{author}{\bibinfo{person}{F. Chicano}, \bibinfo{person}{D. Whitley},
  \bibinfo{person}{G. Ochoa}, {and} \bibinfo{person}{R. Tin{\'o}s}.}
  \bibinfo{year}{2017}\natexlab{}.
\newblock \showarticletitle{Optimizing one million variable NK landscapes by
  hybridizing deterministic recombination and local search}. In
  \bibinfo{booktitle}{\emph{Proceedings of the Genetic and Evolutionary
  Computation Conference}}. ACM, \bibinfo{pages}{753--760}.
\newblock


\bibitem[\protect\citeauthoryear{Dasgupta and Michalewicz}{Dasgupta and
  Michalewicz}{2013}]%
        {dasgupta2013evolutionary}
\bibfield{author}{\bibinfo{person}{Dipankar Dasgupta} {and}
  \bibinfo{person}{Zbigniew Michalewicz}.} \bibinfo{year}{2013}\natexlab{}.
\newblock \bibinfo{booktitle}{\emph{Evolutionary algorithms in engineering
  applications}}.
\newblock \bibinfo{publisher}{Springer Science \& Business Media}.
\newblock


\bibitem[\protect\citeauthoryear{Deb and Myburgh}{Deb and Myburgh}{2016}]%
        {deb2016breaking}
\bibfield{author}{\bibinfo{person}{Kalyanmoy Deb} {and}
  \bibinfo{person}{Christie Myburgh}.} \bibinfo{year}{2016}\natexlab{}.
\newblock \showarticletitle{Breaking the billion-variable barrier in real-world
  optimization using a customized evolutionary algorithm}. In
  \bibinfo{booktitle}{\emph{Proc. GECCO 2016}}. ACM, \bibinfo{pages}{653--660}.
\newblock


\bibitem[\protect\citeauthoryear{Derbel and Canonne}{Derbel and
  Canonne}{2021}]%
        {derbel2021graph}
\bibfield{author}{\bibinfo{person}{Bilel Derbel} {and} \bibinfo{person}{Lorenzo
  Canonne}.} \bibinfo{year}{2021}\natexlab{}.
\newblock \showarticletitle{A graph coloring based parallel hill climber for
  large-scale NK-landscapes}. In \bibinfo{booktitle}{\emph{Proceedings of the
  Genetic and Evolutionary Computation Conference}}. \bibinfo{pages}{216--224}.
\newblock


\bibitem[\protect\citeauthoryear{Gronau and Moran}{Gronau and Moran}{2007}]%
        {gronau2007optimal}
\bibfield{author}{\bibinfo{person}{I. Gronau} {and} \bibinfo{person}{S.
  Moran}.} \bibinfo{year}{2007}\natexlab{}.
\newblock \showarticletitle{Optimal implementations of {UPGMA} and other common
  clustering algorithms}.
\newblock \bibinfo{journal}{\emph{{Information Processing Letters}}}
  \bibinfo{volume}{104}, \bibinfo{number}{6} (\bibinfo{year}{2007}),
  \bibinfo{pages}{205--210}.
\newblock


\bibitem[\protect\citeauthoryear{Jabir, Panicker, and Sridharan}{Jabir
  et~al\mbox{.}}{2015}]%
        {jabir2015multi}
\bibfield{author}{\bibinfo{person}{E Jabir}, \bibinfo{person}{Vinay~V
  Panicker}, {and} \bibinfo{person}{R Sridharan}.}
  \bibinfo{year}{2015}\natexlab{}.
\newblock \showarticletitle{Multi-objective optimization model for a green
  vehicle routing problem}.
\newblock \bibinfo{journal}{\emph{Procedia-Social and Behavioral Sciences}}
  \bibinfo{volume}{189} (\bibinfo{year}{2015}), \bibinfo{pages}{33--39}.
\newblock


\bibitem[\protect\citeauthoryear{Jozefowiez, Semet, and Talbi}{Jozefowiez
  et~al\mbox{.}}{2008}]%
        {jozefowiez2008multi}
\bibfield{author}{\bibinfo{person}{Nicolas Jozefowiez},
  \bibinfo{person}{Fr{\'e}d{\'e}ric Semet}, {and} \bibinfo{person}{El-Ghazali
  Talbi}.} \bibinfo{year}{2008}\natexlab{}.
\newblock \showarticletitle{Multi-objective vehicle routing problems}.
\newblock \bibinfo{journal}{\emph{European journal of operational research}}
  \bibinfo{volume}{189}, \bibinfo{number}{2} (\bibinfo{year}{2008}),
  \bibinfo{pages}{293--309}.
\newblock


\bibitem[\protect\citeauthoryear{Karp}{Karp}{1972}]%
        {karp1972reducibility}
\bibfield{author}{\bibinfo{person}{Richard~M Karp}.}
  \bibinfo{year}{1972}\natexlab{}.
\newblock \showarticletitle{Reducibility among combinatorial problems}.
\newblock In \bibinfo{booktitle}{\emph{Complexity of computer computations}}.
  \bibinfo{publisher}{Springer}, \bibinfo{pages}{85--103}.
\newblock


\bibitem[\protect\citeauthoryear{Li and Yu}{Li and Yu}{2017}]%
        {li2017speeding}
\bibfield{author}{\bibinfo{person}{Sung-Chi Li} {and} \bibinfo{person}{Tian-Li
  Yu}.} \bibinfo{year}{2017}\natexlab{}.
\newblock \showarticletitle{Speeding Up {DSMGA-II} on {CUDA} Platform}. In
  \bibinfo{booktitle}{\emph{Proceedings of the Genetic and Evolutionary
  Computation Conference}} (Berlin, Germany) \emph{(\bibinfo{series}{GECCO
  '17})}. \bibinfo{publisher}{ACM}, \bibinfo{address}{New York, NY, USA},
  \bibinfo{pages}{809--816}.
\newblock
\showISBNx{978-1-4503-4920-8}


\bibitem[\protect\citeauthoryear{Luong, Alderliesten, Bel, Niatsetski, and
  Bosman}{Luong et~al\mbox{.}}{2018}]%
        {luong2018application}
\bibfield{author}{\bibinfo{person}{Ngoc~Hoang Luong}, \bibinfo{person}{Tanja
  Alderliesten}, \bibinfo{person}{Arjan Bel}, \bibinfo{person}{Yury
  Niatsetski}, {and} \bibinfo{person}{Peter A~N Bosman}.}
  \bibinfo{year}{2018}\natexlab{}.
\newblock \showarticletitle{Application and benchmarking of multi-objective
  evolutionary algorithms on high-dose-rate brachytherapy planning for prostate
  cancer treatment}.
\newblock \bibinfo{journal}{\emph{Swarm and Evolutionary Computation}}
  \bibinfo{volume}{40} (\bibinfo{year}{2018}), \bibinfo{pages}{37--52}.
\newblock


\bibitem[\protect\citeauthoryear{Mittal and Vaishay}{Mittal and
  Vaishay}{2019}]%
        {mittal2019survey}
\bibfield{author}{\bibinfo{person}{Sparsh Mittal} {and}
  \bibinfo{person}{Shraiysh Vaishay}.} \bibinfo{year}{2019}\natexlab{}.
\newblock \showarticletitle{A survey of techniques for optimizing deep learning
  on {GPUs}}.
\newblock \bibinfo{journal}{\emph{Journal of Systems Architecture}}
  \bibinfo{volume}{99} (\bibinfo{year}{2019}), \bibinfo{pages}{101635}.
\newblock


\bibitem[\protect\citeauthoryear{{NVIDIA Corporation}}{{NVIDIA
  Corporation}}{2017}]%
        {nvidia2017tesla}
\bibfield{author}{\bibinfo{person}{{NVIDIA Corporation}}.}
  \bibinfo{year}{2017}\natexlab{}.
\newblock \bibinfo{title}{{NVIDIA} {Tesla} {V100} {GPU} {architecture} : {The}
  world’s most advanced data center {GPU}}.
\newblock
\newblock


\bibitem[\protect\citeauthoryear{{NVIDIA Corporation}}{{NVIDIA
  Corporation}}{2018}]%
        {nvidia2010programming}
\bibfield{author}{\bibinfo{person}{{NVIDIA Corporation}}.}
  \bibinfo{year}{2018}\natexlab{}.
\newblock \bibinfo{title}{{CUDA} {C} Programming guide v9.1.85}.
\newblock
\newblock


\bibitem[\protect\citeauthoryear{Rendl, Rinaldi, and Wiegele}{Rendl
  et~al\mbox{.}}{2010}]%
        {rendl2010solving}
\bibfield{author}{\bibinfo{person}{Franz Rendl}, \bibinfo{person}{Giovanni
  Rinaldi}, {and} \bibinfo{person}{Angelika Wiegele}.}
  \bibinfo{year}{2010}\natexlab{}.
\newblock \showarticletitle{Solving max-cut to optimality by intersecting
  semidefinite and polyhedral relaxations}.
\newblock \bibinfo{journal}{\emph{Mathematical Programming}}
  \bibinfo{volume}{121}, \bibinfo{number}{2} (\bibinfo{year}{2010}),
  \bibinfo{pages}{307--335}.
\newblock


\bibitem[\protect\citeauthoryear{Sudholt}{Sudholt}{2015}]%
        {sudholt2015parallel}
\bibfield{author}{\bibinfo{person}{Dirk Sudholt}.}
  \bibinfo{year}{2015}\natexlab{}.
\newblock \showarticletitle{Parallel evolutionary algorithms}.
\newblock In \bibinfo{booktitle}{\emph{Springer Handbook of Computational
  Intelligence}}. \bibinfo{publisher}{Springer}, \bibinfo{pages}{929--959}.
\newblock


\bibitem[\protect\citeauthoryear{Thierens and Bosman}{Thierens and
  Bosman}{2012}]%
        {thierens2012predetermined}
\bibfield{author}{\bibinfo{person}{Dirk Thierens} {and} \bibinfo{person}{Peter
  Bosman}.} \bibinfo{year}{2012}\natexlab{}.
\newblock \showarticletitle{Predetermined versus learned linkage models}. In
  \bibinfo{booktitle}{\emph{Proc. GECCO.}} ACM, \bibinfo{pages}{289--296}.
\newblock


\bibitem[\protect\citeauthoryear{Thierens and Bosman}{Thierens and
  Bosman}{2011}]%
        {thierens2011optimal}
\bibfield{author}{\bibinfo{person}{D. Thierens} {and} \bibinfo{person}{P.~A.~N.
  Bosman}.} \bibinfo{year}{2011}\natexlab{}.
\newblock \showarticletitle{Optimal mixing evolutionary algorithms}. In
  \bibinfo{booktitle}{\emph{Proc. GECCO}}. ACM, \bibinfo{pages}{617--624}.
\newblock


\bibitem[\protect\citeauthoryear{Tintos, Whitley, and Chicano}{Tintos
  et~al\mbox{.}}{2015}]%
        {tintos2015partition}
\bibfield{author}{\bibinfo{person}{R. Tintos}, \bibinfo{person}{D. Whitley},
  {and} \bibinfo{person}{F. Chicano}.} \bibinfo{year}{2015}\natexlab{}.
\newblock \showarticletitle{Partition crossover for pseudo-boolean
  optimization}. In \bibinfo{booktitle}{\emph{Proceedings of the 2015 ACM
  Conference on Foundations of Genetic Algorithms XIII}}. ACM,
  \bibinfo{pages}{137--149}.
\newblock


\bibitem[\protect\citeauthoryear{Welsh and Powell}{Welsh and Powell}{1967}]%
        {welsh1967upper}
\bibfield{author}{\bibinfo{person}{Dominic J~A Welsh} {and}
  \bibinfo{person}{Martin~B Powell}.} \bibinfo{year}{1967}\natexlab{}.
\newblock \showarticletitle{An upper bound for the chromatic number of a graph
  and its application to timetabling problems}.
\newblock \bibinfo{journal}{\emph{Comput. J.}} \bibinfo{volume}{10},
  \bibinfo{number}{1} (\bibinfo{year}{1967}), \bibinfo{pages}{85--86}.
\newblock


\bibitem[\protect\citeauthoryear{Wong, Wong, and Fok}{Wong
  et~al\mbox{.}}{2005}]%
        {wong2005parallel}
\bibfield{author}{\bibinfo{person}{Man-Leung Wong}, \bibinfo{person}{Tien-Tsin
  Wong}, {and} \bibinfo{person}{Ka-Ling Fok}.} \bibinfo{year}{2005}\natexlab{}.
\newblock \showarticletitle{Parallel evolutionary algorithms on graphics
  processing unit}. In \bibinfo{booktitle}{\emph{2005 IEEE Congress on
  Evolutionary Computation}}, Vol.~\bibinfo{volume}{3}. IEEE,
  \bibinfo{pages}{2286--2293}.
\newblock


\end{thebibliography}

\end{document}